\pdfoutput=1
\documentclass[11pt]{article}

\usepackage[preprint]{acl}

\usepackage{times}
\usepackage{latexsym}
 \usepackage{ulem}
\usepackage[T1]{fontenc}
\usepackage[utf8]{inputenc}

\usepackage{microtype}
\usepackage{inconsolata}
\usepackage{microtype}
\usepackage{booktabs} % for professional tables
\usepackage{kotex}%Including images in your LaTeX document requires adding
\usepackage{hyperref}
\usepackage[labelformat=simple]{subcaption}

\usepackage{graphicx}
\usepackage{algorithm}
\usepackage{algorithmic}

\usepackage{amsmath}
\usepackage{amssymb}
\usepackage{mathtools}
\usepackage{amsthm}
\usepackage{multirow}
\usepackage{makecell}
\usepackage{array}
\usepackage{longtable}
\usepackage{listings}
\usepackage{xcolor}
\usepackage{caption}
\usepackage{tabularx}
\usepackage{tikz}
\def\checkmark{\tikz\fill[scale=0.4](0,.35) -- (.25,0) -- (1,.7) -- (.25,.15) -- cycle;} 
\newcommand{\scv}{SC-Energy}

\title{Introducing Verification Task of Set Consistency with \\Set-Consistency Energy Networks}

\author{Mooho Song, Hyeryung Son, Jay-Yoon Lee*\\
        Seoul National University \\ \{anmh9161, hyeryung.son, lee.jayyoon\}@snu.ac.kr}

\begin{document}
\maketitle

% \begin{abstract}

% Examining logical inconsistencies among multiple statements is a crucial challenge in machine learning, particularly when dealing with collection of sentences or question-answer pairs. Traditional methods often rely on pairwise comparisons, which fail to capture inconsistencies that only emerge when multiple statements are considered collectively. To overcome these challenges, we propose the \textit{Set-Consistency Energy Network} (\scv), a novel approach designed to examine logical inconsistency across all the statements in a set.
% \scv~employs a contrastive loss framework to learn the compatibility among a collection of statements, whether in the form of sentences or question-answer pairs. 
% \scv~ significantly outperforms existing approaches, including prompting-based LLM models, in efficiently verifying inconsistencies and identifying the specific statements that contribute to logical contradictions. We also introduce the task of set-consistency verification, which extends natural language inference (NLI) beyond pairwise comparisons to assess logical inconsistency among multiple statements. We release two new datasets—Set-LConVQA and Set-SNLI—tailored for set-consistency verification tasks. \end{abstract}

\begin{abstract}
Examining logical inconsistencies among multiple statements—such as collections of sentences or question-answer pairs—is a crucial challenge in machine learning, particularly for ensuring the safety and reliability of models. Traditional methods that rely on pairwise comparisons often fail to capture inconsistencies that only emerge when more than two statements are evaluated collectively. To address this gap, we introduce the task of \textit{set-consistency verification}, an extension of natural language inference (NLI) that assesses the logical coherence of entire sets rather than isolated pairs. Building on this task, we present the \textit{Set-Consistency Energy Network} (\scv), a novel model that employs a contrastive loss framework to learn the compatibility among a collection of statements. Our approach not only efficiently verifies inconsistencies and pinpoints the specific statements responsible for logical contradictions, but also significantly outperforms existing methods—including prompting-based LLM models. Furthermore, we release two new datasets: Set-LConVQA and Set-SNLI for set-consistency verification task.
\end{abstract}

\section{Introduction}
In machine learning (ML), examining whether multiple statements exhibit logical consistency is a crucial challenge, especially when evaluating models for their safe usage.
Examples from document summarization and question answering provide intuitive reason for such evaluations. When a large language model (LLM) generates a summary of a document, inconsistencies may arise between the original document and the summary. Similarly, in question-answering tasks, ML models do not inherently guarantee consistent responses to semantically related questions. 
Consider the following question-answering example from \cite{tandon2019wiqa}. If a model answers “Yes” to the question “Does CO$_2$ increase the population of polar bears?”, it should not also answer “Yes” to the contradictory question “Does CO$_2$ decrease the population of polar bears?”.
Logical consistency among these pairs, and more broadly within a set, is required for reliability and underscores the need for effective inconsistency detection methods.

Few studies have addressed related problems, particularly in the domain of factual inconsistency detection. 
These studies primarily focus on identifying factual discrepancies between a document and its summary. 
However, detecting logical inconsistencies among multiple statements remains limited with these methods. 
% However, existing approaches have limitations that make it difficult to directly extend them to verifying logical inconsistencies among multiple statements.

%%% previous writing
% Although few studies have addressed this exact problem, related work in factual inconsistency detection has tackled similar issues. These studies primarily focus on identifying factual discrepancies between a document and its summary. However, existing approaches have limitations that make it difficult to directly extend them to verifying logical inconsistencies among multiple statements.

First, some studies do not incorporate a holistic view of consistency across multiple statements. For example, \cite{fid-xu2024identifying,fid-wang-etal-2020-asking,fid-fabbri-etal-2022-qafacteval} generate questions based on summaries and then obtain answers from the original document and summaries simultaneously, assessing consistency between them. However, arbitrary question generations do not guarantee the detection of all subtle logical inconsistencies, making it challenging to capture a comprehensive view of consistency for multiple statements.

Second, even approaches that aim for a holistic view face limitations when verifying consistency among multiple statements. \cite{fid-yang-etal-2024-reassess,fid-falke-etal-2019-ranking} evaluate factual consistency based on (NLI), typically by comparing the entailment score for all combinations of pairs of sentences (or chunk of sentences) present in a document and its summary. Although this pairwise comparing method considers all sentences in a document, it has inherent shortcomings as following:

\begin{list}{\labelitemi}{\leftmargin=1em}
    \item \textbf{Limited Scope of Inconsistency Detection:} Existing Method can only detect inconsistencies when two statements directly conflict each other. In other words, logical inconsistencies that emerge only when multiple statements are considered collectively may go undetected. For example, consider the following three statements:
    \begin{itemize}
        \item \textit{"Either the train arrives at 8 AM or it arrives at 9 AM."} \textcolor{gray}{($p$ or $q$)}
        \item \textit{"The train does not arrive at 8 AM."} \textcolor{gray}{($\neg p$)}
        \item \textit{"The train does not arrive at 9 AM."} \textcolor{gray}{($\neg q$)}
    \end{itemize}
    While no two statements contradict each other, an inconsistency becomes evident when all three are considered together.
    
    \item \textbf{Combinatorial Complexity}: In large text corpora such as articles, combinatorial complexity of pairwise comparisons across all statements incur heavy computational costs.
    
    \item \textbf{Overly Sensitive Classification:} When examining inconsistencies in exhaustive pairwise comparisons, if even one of them detects an inconsistency, the entire document is deemed inconsistent in a naive way. As the number of statements increases, the probability of detecting at least one inconsistency also increases, 
    rendering pairwise classifications impractical for a larger set.
    % rendering binary classification impractical for longer texts.

\end{list}

To overcome these limitations, we propose the \textbf{Set-Consistency Energy Network} (\scv), an approach that detects the logical inconsistencies across the entire statements within a set.
\scv~ treats a collection of natural language statements - sentences or input-output pairs - as a set and employs a contrastive loss framework to learn the compatibility among them.
Unlike traditional research related to energy networks taht process single inputs \cite{belanger2016structured,lecun2006tutorial,tu2019improving-infnet,lee2022structured}
, \scv~ is designed to capture the compatibility across multiple data points.
Even with a relatively compact architecture such as RoBERTa-base \cite{liu2019roberta}, out model significantly outperforms LLMs such as GPT-4o \cite{achiam2023gpt} in effectively detecting inconsistencies.

Our key contributions are as follows:
\begin{list}{\labelitemi}{\leftmargin=1em}
    \item We introduce the task of \textit{set-consistency verification}, which extends natural language inference beyond pairwise comparisons to assess the logical coherence of multiple statements. We show that LLMs lack such verification capability and highlight the importance of the task. 
    \item We release two refactored datasets, Set-LConVQA and Set-SNLI, for set-consistency verification and locate tasks to facilitate further research in this area.
    \item We show that learning an energy space capable of distinguishing subtle differences in consistency across entire sets of statements, rather than binary classification of an isolated set, is crucial. This contrast is particularly highlighted in the Locate task, which requires pinpointing the specific statements responsible for logical contradictions.
    % In contrast, relying solely on single binary classification on isolated pairs yields poor performance at pinpointing the specific statements responsible for logical contradictions.
\end{list}

% Extensive experiments demonstrate that \scv~significantly outperforms existing approaches in detecting inconsistencies and identifying the contributing statements, paving the way for safer and more reliable ML applications.

\section{Related Works}

\paragraph{Factual Inconsistency Detection}
Several studies have focused on identifying factual discrepancies in summarization tasks. In particular, research such as \cite{fid-luo2023chatgpt, fid-xu2024identifying, fid-yang-etal-2024-reassess, fid-fabbri-etal-2022-qafacteval, fid-laban-etal-2022-summac} has examined methods to detect inconsistencies between a document and its summary. These approaches typically involve decomposing the text into smaller units, generating common questions to compare source context and summary, or performing pairwise entailment-score comparisons for all combinations of individual sentences (or, chunk of sentences) from document and summary.
% However, while effective to a degree, QA-based methods can introduce additional uncertainty due to model variability, and they may not fully capture the holistic consistency across the entire document.

\paragraph{Structured Energy Networks}
Several studies have explored energy-based models for structured prediction tasks. Belanger and McCallum (2016) introduced Structured Prediction Energy Networks (SPENs), which use energy functions to model dependencies between output components in structured tasks \cite{belanger2016structured}. Tu et al. (2019) enhanced the joint training of inference networks and SPENs with a compound objective, leading to more effective optimization \cite{tu2019improving-infnet}. Lee et al. (2022) proposed SEAL, a framework where an energy network is used as a loss function to guide the training of a separate neural network \cite{lee2022structured}.

A characteristic feature of these studies is that they perform contrastive learning over points in the $(\mathcal{X} \times \mathcal{Y})$ space to train the energy network. Our work expands this concept by learning the energy surface in the $(\mathcal{X} \times \mathcal{Y})^{*}$ space—i.e., over arbitrary number of input-output pairs (as discussed in Section~\ref{sec:method:\scv}).

\section{Set-Consistency Energy Networks}
\label{sec:method:\scv}

\subsection{Definitions}
\label{sec:definitions}

\begin{table*}[h]
\begin{center}
\scalebox{0.75}{\begin{tabular}{cll}
\toprule
Data Consistency & \makecell[c]{Set-LConVQA\\Example} & \makecell[c]{Set-SNLI\\Example}\\
\midrule
Consistent Set ($S_{C}$) &  \makecell[l]{\{("question": "what color is desk?", "answer": "brown"),\\("question": "is desk brown?", "answer": "yes"),\\("question": "is desk pink?", "answer": "\textbf{no}")\}} & \makecell[l]{\{"If a couple walk hand in hand down a street, \\then a couple is   walking together.", \textcolor{gray}{($p \rightarrow q$)}\\      "No couple is walking together.", \textcolor{gray}{($\neg q$)}\\      "\textbf{No couple walks hand in hand down a street.}" \textcolor{gray}{($\neg p$)}\}}\\
\cmidrule{1-3}
Inconsistent Set ($S_{I}$) &  \makecell[l]{\{("question": "what color is desk?", "answer": "brown"),\\("question": "is desk brown?", "answer": "yes"),\\("question": "is desk pink?", "answer": "\textbf{yes}")\}} & \makecell[l]{\{"If a couple walk hand in hand down a street, \\then a couple is   walking together.", \textcolor{gray}{($p \rightarrow q$)}\\      "No couple is walking together.", \textcolor{gray}{($\neg q$)}\\      "\textbf{Either a couple walk hand in hand down a street,}\\\textbf{or a couple is walking together.}"\textcolor{gray}{($p \vee q$)}\}}\\
\bottomrule
\end{tabular}}
\end{center}
\caption{Examples of consistent ($S_{C}$) and inconsistent ($S_{I}$) sets from the Set-LConVQA and Set-SNLI datasets. The bolded portions indicate the different statements that distinguish the consistent set from the inconsistent set. The gray-colored propositional symbols explain the logical relationship in sentences. In the Set-LConVQA example, inconsistency is explicitly evident in the QA pairs (e.g., conflicting answers about the desk's color), whereas the Set-SNLI example illustrates more subtle logical discrepancies among sentences as it emerges from nuanced differences in phrasing and logical entailment rather than explicit factual contradictions.}

\label{tab:set-lconvqa-snli-examples}
\end{table*}

Several definitions are listed as follows.

A \textbf{statement} ($s$) refers to an individual unit of information, such as a sentence / sentences in a news article or a QA-pair in a question-answering dataset.

\textbf{Set} ($S$) refers to a collection of statements. The size of a set $S$, denoted as $|S|$, represents the number of statements within $S$. In this paper, we consider only finite sets.

If there exists a subset $\Tilde{S} \subseteq S$ with $|\Tilde{S}| \geq 2$ such that the statements in $\Tilde{S}$ are logically inconsistent, then $S$ is called an \textbf{inconsistent set}. Otherwise, $S$ is referred to as a \textbf{consistent set}.

% By definition, if a set $S$ is inconsistent, then any superset $T \supseteq S$ is also inconsistent. Conversely, if a set $S$ is consistent, then any subset $U \subseteq S$ is consistent as well.

\subsection{Notations}
\label{sec:notations}
A statement $s$ can be either a standalone sentence or a pair $(x, y)$ (e.g., a QA-pair). We now present the notations for both cases.

\textbf{Standalone Sentence Formulation:}  
Let $\mathcal{S}$ denote the space of standalone sentences. In this formulation, the \scv~is defined as
\[
E_{\theta}: \mathcal{S}^{*} \rightarrow \mathbb{R},
\]
which is a parameterized function that takes an arbitrary number of sentences as input and produces a real-valued energy score.
Note that, the asterisk ($^*$) indicates that the function accepts a sequence (or set) of sentences drawn from $\mathcal{S}$—that is, an arbitrary number of sentences can be provided as input.

\textbf{QA-Pair Formulation:}  
Let $\mathcal{X}$ and $\mathcal{Y}$ denote the input and output spaces, respectively. For statements represented as QA-pairs, the \scv~is defined as
\[
E_{\theta}: (\mathcal{X} \times \mathcal{Y})^{*} \rightarrow \mathbb{R}.
\]
In this setting, the network processes an arbitrary number of $(x, y)$ pairs and outputs a real-valued energy score.

For simplicity, we denote a consistent set as $S_C$ and an inconsistent set as $S_I$. When constructing new sets by merging different sets, we concatenate their indices to indicate their composition. For instance, the union of two distinct consistent sets is denoted as $S_{CC}$. We consider the union of different consistent sets to remain consistent—thus, $S_{CC}$ is essentially a consistent set and could be regarded as $S_C$. However, for more fine-grained analysis, rather than simplifying it with $S_{C}$, we explicitly denote the data generation process and label such set as $S_{CC}$ in our experiments.

\subsection{Training \scv}
The function $E_{\theta}$ is trained to assign lower energy values to consistent sets and higher energy values to inconsistent sets. Given a consistent set $S_C$ and an inconsistent set $S_I$, the loss function $L_E$ is defined as:
\begin{align}
\nonumber
    L_{E}(S_{C}, S_{I}) = \left[ E_{\theta}(S_{C}) - E_{\theta}(S_{I}) + \alpha \right]_{+},
\end{align}
where $[\cdot]_{+} = \max(\cdot, 0)$ and $\alpha$ is a hyperparameter.

In Section \ref{sec:constructing sc si}, we discuss the methods for constructing $S_C$ and $S_I$; detailed procedures for each dataset are provided in Section \ref{sec:datasets}. In Section \ref{sec:fine-grained training \scv}, we introduce the training methodology for \scv~using $S_C$ and $S_I$.

\subsubsection{Construction of $S_{C}$ and $S_{I}$}
\label{sec:constructing sc si}
We first introduce construction of $S_{C}$ and $S_{C}$ based on QA-pair setting. Given a consistent set 
\(
S_C = \{(x_1, y_1), (x_2, y_2), \dots, (x_n, y_n)\},
\)
we construct an inconsistent set $S_I$ such that it preserves the overall semantic content and form of $S_C$ while introducing logical inconsistencies. The simplest example is to modify one output $y_i$ in a QA-pair by replacing it with $y^{*}_{i}$, thereby yielding
\(
S_I = \{(x_1, y_1), \dots, (x_i, y^{*}_{i}), \dots, (x_n, y_n)\}.
\)
In this construction, $S_I$ remains similar to $S_C$ in terms of semantics and structure, yet the altered element induces a logical inconsistency. In the context of a set of sentences, a similar strategy can be adopted by prompting an LLM to regenerate some of the sentences in $S_C$ to produce $S_I$. 

These constructions can be carried out using purely rule-based methods (e.g., Set-LConVQA) or through a combination of rule-based techniques and LLM-generated modifications (e.g., Set-SNLI), which will be introduced in section \ref{sec:datasets} in detail.

\subsubsection{Fine-grained Training of \scv}
\label{sec:fine-grained training \scv}
We derive eight contrastive signals to train $E_{\theta}$, capturing varying levels of logical consistency. Each of these eight contrasts contributes a loss term that guides the model to assign lower energy to more consistent sets and higher energy to more inconsistent ones. The loss function for \scv~incorporates the $L_E$ values from all eight contrasts:

\begin{enumerate}
    \item \textbf{Basic Contrast:} 
    \begin{list}{\labelitemi}{\leftmargin=1em}

        \item ($S_C$ vs. $S_I$): A direct comparison between a consistent set and an inconsistent set.
    \end{list}
    \item \textbf{Union-Based Contrasts:} By forming unions, we generate an additional consistent set $S_{CC}$ and two extra inconsistent sets $S_{CI}$ and $S_{II}$. This yields five additional comparisons:
    \begin{list}{\labelitemi}{\leftmargin=1em}
    \item ($S_C$ vs. $S_{CI}$), ($S_C$ vs. $S_{II}$), ($S_{CC}$ vs. $S_I$), \\ ($S_{CC}$ vs. $S_{CI}$), ($S_{CC}$ vs. $S_{II}$).
    \end{list}
    % \begin{list}{\labelitemi}{\leftmargin=1em}
    %     \item $S_C$ vs. $S_{CI}$
    %     \item $S_C$ vs. $S_{II}$
    %     \item $S_{CC}$ vs. $S_I$
    %     \item $S_{CC}$ vs. $S_{CI}$
    %     \item $S_{CC}$ vs. $S_{II}$
    % \end{list}
    \item \textbf{Inconsistency Degree Contrasts}
    \begin{list}{\labelitemi}{\leftmargin=1em}
    \item \textbf{($S_{CI}$ vs. $S_I$):} In $S_{CI}$, we regard the presence of additional neutral (or consistent) elements alongside inconsistent ones makes it less contradictory overall than the fully inconsistent $S_I$. Thus, we regard $S_{CI}$ to be more consistent (i.e., of lower inconsistency degree) than $S_I$.
    \item \textbf{($S_I$ vs. $S_{II}$):} Here, $S_{II}$ is formed by uniting multiple inconsistent sets, thereby amplifying the overall contradiction. Consequently, we treat $S_{II}$ to have higher degree of inconsistency than $S_I$.
    \end{list}
\end{enumerate}

\subsection{Inference and Thresholding}
Since $E_{\theta}$ is a real-valued function, a predefined threshold is used to determine set consistency. A set is classified as consistent if its energy score is below the threshold and inconsistent otherwise. 
\newline

Detailed information on the training procedure using the eight contrasts, the conversion of a set $S$ into the input for \scv, and the selection of the threshold can be found in appendix~\ref{appendix: training inference detail of energy network, threshold learning}.

\section{Datasets}
\label{sec:datasets}

To evaluate the proposed task of set-consistency verification, we introduce two new datasets: Set-LConVQA and Set-SNLI. Set-LConVQA is derived from the existing LConVQA dataset \cite{ray-etal-2019-sunny}, and Set-SNLI is constructed by modifying the SNLI dataset \cite{bowman-etal-2015-large}. Both datasets are designed to group multiple instances into logically consistent or inconsistent sets, thereby enabling systematic evaluation of set-consistency verification.

We provide four types of data splits for both datasets: training, validation1, validation2, and test sets. The validation1, validation2, and test sets contain 200 consistent and 200 inconsistent sets each. The training sets contain 6,754 and 6,225 consistent and inconsistent sets for Set-LConVQA and Set-SNLI, respectively. Below, we describe each dataset in detail.

\subsection{Set-LConVQA}
\label{sec:dataset:set-lconvqa}
Set-LConVQA is designed to assess whether a given set of question-answer (QA) pairs is mutually consistent or inconsistent. The original LConVQA dataset is a Visual Question Answering (VQA) dataset automatically generated from the Visual Genome scene graph by extracting object, attribute, and relation information to create logically consistent / inconsistent QA pairs. Set-LConVQA removes the image dependency and focuses purely on the textual QA pairs.

An example from the dataset is shown in Table~\ref{tab:set-lconvqa-snli-examples}. A key characteristic of Set-LConVQA is that in each inconsistent set $S_I$, there exists a specific pair of QA pairs, denoted as $s_i$ and $s_j$, such that the size-2 set $\{s_i, s_j\}$ is inconsistent. This property distinguishes Set-LConVQA from the Set-SNLI dataset, which is introduced in Section~\ref{sec:dataset:set-snli}. Moreover, through manual verification, we confirmed that in all test instances where the set size is at least four, there is exactly one element in $S_I$ whose removal renders the set consistent. Therefore, in addition to set-consistency verification, we can  utilize Set-LConVQA for an additional task (see Section~\ref{sec:experiment: locate main}) to show the usefulness of \scv. 
Further details regarding the Set-LConVQA dataset are provided in Appendix~\ref{appendix: set lconvqa details}.

\subsection{Set-SNLI}
\label{sec:dataset:set-snli}
Set-SNLI dataset can be used to evaluate the ability to determine whether a set of natural language sentences is logically consistent. Unlike traditional Natural Language Inference (NLI) tasks that assess relationships between a single premise and a hypothesis, Set-SNLI requires reasoning over multiple sentences to capture more complex logical interactions.

For example, consider the three sentences:
    ``Either the train arrives at 8 AM or it arrives at 9 AM.'',
     ``The train does not arrive at 8 AM.'',
 ``The train does not arrive at 9 AM.''.
No two sentences contradict each other when examined pairwise, but collectively they introduce a contradiction. Since the ordering of sentences is irrelevant in this task, traditional entailment relationships in NLI cannot be determined. Instead, sets are classified as either consistent or inconsistent, with inconsistency implying that not all sentences in the set can simultaneously be true.

Inspired by \cite{nakamura-etal-2023-logicattack}, we construct Set-SNLI by transforming SNLI sentence pairs into multi-sentence sets using predefined logical rules. Examples are presented in Table~\ref{tab:set-lconvqa-snli-examples}, and detailed construction steps are provided in Appendix~\ref{appendix:set-snli_dataset}.

In contrast to Set-LConVQA, which guarantees that for each inconsistent set $S_I$ there always exists a size-2 subset $\tilde{S} \subset S$ that is inconsistent, this property is not assured in the Set-SNLI dataset. In Set-SNLI, some inconsistent sets exhibit this property while others do not. 

Set-SNLI is more challenging than Set-LConVQA because the inconsistencies in Set-SNLI tend to be more subtle and require holistic reasoning over multiple sentences. While Set-LConVQA usually presents explicit factual contradictions (e.g., conflicting answers about a desk's color), the logical discrepancies in Set-SNLI arise from nuanced differences in phrasing and entailment among sentences, making it harder to detect inconsistencies.

\section{Experiments}
\label{sec:Experiments}

This section presents the experimental results of set-consistency verification and downstream task, demonstrating that \scv~outperforms other baselines. For performance comparison, the baselines are evaluated along two axes: \textbf{Model Architecture} and \textbf{Verification Strategy}.

\textbf{Model Architecture}: This axis categorizes the baseline models based on their output representations and training objectives. In particular, the differences lie in the output format: LLM-based models output natural language, binary classifiers output a 2-dimensional vector, and energy-based models output a single real-valued score. These models can be divided into three types:
\begin{enumerate}
    \item \textbf{LLM-based}: Utilizes a pre-trained, frozen large language model to assess set consistency via prompt-based querying. We employed the GPT-4o model from OpenAI\footnote{https://openai.com/}, with Chain-of-Thought prompting \cite{wei2022chain}.
    \item \textbf{Binary Classifier (2-dim vector)}: A conventional classification model with an output dimension corresponding to the number of labels (consistent/inconsistent), trained using a cross-entropy loss function.
    \item \textbf{Energy-based (Real-valued)}: Following energy-based model's principle, this model produces a continuous real-valued output, trained such that consistent sets yield lower scores and inconsistent sets yield higher scores. Training is performed using a contrastive loss.
\end{enumerate}

\textbf{Verification strategy}: This axis defines how consistency is evaluated within a set: 
\begin{enumerate}
    \item \textbf{Element-wise Verification}: Similar to \cite{fid-yang-etal-2024-reassess,fid-falke-etal-2019-ranking}, this strategy evaluates logical inconsistency by comparing all possible pairs of statements within a set. For a set $S$ of size $|S| = N$, this requires $\frac{N(N-1)}{2}$ pairwise comparisons, where each pair is assessed for 1:1 consistency.
    \item \textbf{Set-level Verification}: Directly inputs the entire set into the model, reducing computational overhead and capturing inconsistencies that may be overlooked by pairwise comparisons.
\end{enumerate}
% \textcolor{red}{\textbf{[미완성인 부분]} Pairwise와 set-level에서 데이터를 어떻게 구성하여 학습/평가하는지에 대한 차이점은 섹션 \ref{sec:datasets}, dataset 설명에서 자세히 다룰 예정입니다.}

To clarify, our \scv~model employs a energy-based (real-valued) model architecture along with a set-level verification strategy. The training and evaluation procedures vary slightly depending on the model architecture and verification strategy used. For detailed information on the prompts and experimental methods, please refer to Appendix~\ref{appendix: set-consistency verification experiment setting detail}.

\subsection{Set-Consistency Verification}
\label{sec:set-consistency verification main section}

Set-consistency verification determines whether a given set $S$ is logically consistent or inconsistent. Beyond evaluating individual sets such as $S_{C}$ and $S_{I}$, the consistency assessment extends to unions of multiple sets. 
% Training is conducted as described in Section \ref{sec:fine-grained training \scv}. 
To evaluate set-consistency verification performance, we generate diverse types of sets to assess the model’s ability to generalize to previously unseen set configurations. The evaluation data includes not only $S_{C}$ and $S_{I}$ but also sets constructed by merging two, three, or four different sets, thereby generating up to 14 possible dataset combinations with different labels (for example, $S_{C}$, $S_{I}$, $S_{CC}$, $\cdots$, $S_{IIII}$).

Table \ref{tab:parta} presents the Macro-F1 score results for set consistency-verification tasks across various model architectures and verification strategies. The average and standard deviation for 5 times of different seeds are provided. Since both the binary classifier and energy-based architectures are data-driven, the specific training datasets used for each model are indicated by check marks (\checkmark) in the Training Data column. For example, if both Set-LConVQA and Set-SNLI are checked, it implies that the model was trained on both datasets simultaneously. 

Regarding the \textit{verification strategies}, the set-level verification strategy consistently outperforms the element-wise strategy across all model architectures, highlighting the importance of providing the entire set as input for verifying logical inconsistencies. Note that, the element-wise verification strategy tends to predict a set as inconsistent with high probability. In fact, when the model uniformly classifies all sets as either consistent or inconsistent, the corresponding Macro-F1 scores are only 0.222 and 0.416, respectively. These results suggest that the element-wise strategy does not achieve sufficiently high performance in set consistency-verification.

% Notably, the frequently observed scores of 0.416 (particularly in the element-wise verification strategy) correspond to cases where the model fails to learn effectively, resulting in predictions that classify all sets as inconsistent. 

% In terms of \textit{model architecture}, the energy-based (real-valued) model outperforms the binary classifier (2-dim vector) model. Furthermore, \scv—which employs the set-level verification strategy combined with the energy-based model architecture—surpasses LLM-based model architectures.

As mentioned earlier, in a element-wise verification approach, if any element-wise verification identifies an inconsistency, the entire document is theoretically deemed inconsistent. This raises the question: to what extent should a certain level of inconsistency be tolerated (even if this is not the ideal approach) in order to achieve optimal performance in set-consistency verification? Further details on this matter can be found in appendix \ref{appendix: section maximum tolerance rate}.

For models in set-level verification strategy, the binary-classifier model architecture exhibits performance comparable to the energy-based model \scv. However, because \scv~learns fine-grained consistency across multiple sets, it performs well on a broader range of tasks beyond set-consistency verification. For further details, please refer to Section \ref{sec:experiment: locate main}.

\subsection{Locating Inconsistent Statements}
\label{sec:experiment: locate main}

% \begin{table}
%     \centering
%     \scalebox{0.75}{\begin{tabular}{ccccc}
%         \toprule
%           & EM & Precision& Recall& F1\\
%           \midrule
%          LLM & 0.734 & 0.846 & \textbf{0.985} & 0.875 \\
%           \midrule
%           \makecell[c]{\textbf{Set-level} comparison \&\\\textbf{Binary Classifier}\\model \\ \jy{Binary Classifier}}  &  0.469 & 0.662 & 0.860 & 0.712\\
%           \midrule
%          \makecell[c]{\scv~(Ours)} & \textbf{0.939} & \textbf{0.967} & \textbf{0.985} & \textbf{0.967}\\
%           % \midrule
%          % \makecell[c]{\scv\\(LConVQA + SNLI)}& 0.808 &0.901 &0.983 & 0.924\\
%          \bottomrule
%     \end{tabular}}
%     \caption{Performance evaluation for locating inconsistent statements within a set. The table compares Exact Match (EM), Precision, Recall, and F1-score across different models. Except for GPT, all of models are trained only on Set-LConVQA dataset. \jy{summary를 caption에}}
%     \label{tab:partb}
% \end{table}

\begin{table}[h]
\centering
\subfloat[Macro-F1 scores for set-consistency verification]{
\scalebox{0.6}{%
\begin{tabular}{ccccll}
\toprule
\multirow{2}{*}{\makecell[l]{Verification\\Strategy}} & \multirow{2}{*}{\makecell[l]{Model\\Architecture}} & \multicolumn{2}{c}{Training Data} & \multicolumn{2}{c}{Test Data} \\
  &  & \makecell[c]{Set-\\LConVQA} & \makecell[c]{Set-\\SNLI} & \makecell[c]{Set-\\LConVQA} & \makecell[c]{Set-\\SNLI}\\
\midrule
\multirow{7}{*}{Element-wise} 
  & LLM & - & - & 0.566 & 0.493 \\
\cmidrule{3-6}
  & \multirow{3}{*}{\makecell[c]{Binary\\Classifier}} 
  & \checkmark &  & 0.577$_{\pm 0.096}$& 0.475$_{\pm 0.055}$ \\ 
  &  &  & \checkmark & 0.434$_{\pm 0.040}$
  &0.490$_{\pm 0.016}$\\ 
  &  & \checkmark & \checkmark & 0.607$_{\pm 0.104}$
  & 0.556$_{\pm 0.022}$ \\
\cmidrule{3-6}
  & \multirow{3}{*}{\makecell[c]{Energy-Based}} 
  & \checkmark &  & 0.697$_{\pm 0.097}$ & 0.523$_{\pm 0.032}$ \\ 
  &  &  & \checkmark & 0.413$_{\pm 0.017}$ & 0.563$_{\pm 0.034}$\\ 
  &  & \checkmark & \checkmark & 0.440$_{\pm 0.003}$ & 0.534$_{\pm 0.027}$\\
\midrule
\multirow{7}{*}{Set-level} 
  & LLM & - & - & 0.926 & 0.710\\
\cmidrule{3-6}
  & \multirow{3}{*}{\makecell[c]{Binary\\Classifier}} 
  & \checkmark &  & 0.985$_{\pm 0.005}$ & 0.474$_{\pm 0.048}$\\ 
  &  &  & \checkmark & 0.533$_{\pm 0.050}$ & 0.973$_{\pm 0.010}$\\ 
  &  & \checkmark & \checkmark & 0.913$_{\pm 0.099}$ & 0.967$_{\pm 0.011}$\\
\cmidrule{3-6}
  & \multirow{3}{*}{\makecell[c]{Energy-Based\\(\scv)}} 
  & \checkmark &  & \textbf{0.987}$_{\pm 0.005}$ & 0.432$_{\pm 0.023}$\\ 
  &  &  & \checkmark & 0.508$_{\pm 0.063}$ & \textbf{0.981}$_{\pm 0.008}$\\ 
  &  & \checkmark & \checkmark & 0.969$_{\pm 0.018}$ & 0.941$_{\pm 0.023}$\\
\bottomrule
\end{tabular}}
\label{tab:parta}
}

\vspace{1em}

\subfloat[Locate Task Performance]{
\scalebox{0.65}{%
\begin{tabular}{ccccc}
\toprule
             Model   & EM   & Precision & Recall    & F1     \\
\midrule
LLM             & 0.734 & 0.846     & {0.985} & 0.875 \\
\midrule
\makecell[c]{Binary Classifier}  
                & 0.784$_{\pm 0.142}$ & 0.880$_{\pm 0.102}$     & \textbf{0.993$_{\pm 0.001}$}
                & 0.910$_{\pm 0.079}$ \\
\midrule
\makecell[c]{Energy-based\\(\scv)} 
                & \textbf{0.923$_{\pm 0.059}$} & \textbf{0.957$_{\pm 0.043}$} & {0.981$_{\pm 0.015}$} & \textbf{0.961$_{\pm 0.035}$} \\
\bottomrule
\end{tabular}}
\label{tab:partb}
}

\caption{(a) Macro-F1 scores for set-consistency verification across different verification strategies and model architectures, and (b) performance evaluation for the Locate task in identifying inconsistent QA pairs. The best performance is highlighted in bold. For both verification (a) and Locate (b), \scv~ shows the best performance for all of datasets. \scv~ is superior at Locate (b) task thanks to learning fine-grained degrees of inconsistencies across diverse sets. Particularly, note that LLMs display poor performance at Set-SNLI, indicating the need for separate training of set verification. For both (a) and (b), average and standard deviation for five different seed pairs are provided.
}
\label{tab:combined}
\end{table}

To showcase the diverse applications of the models trained in Section~\ref{sec:set-consistency verification main section}, we introduce an additional task of \textit{Locate} which evaluate the ability to detect inconsistent statements within a set. We utilize Set-LConVQA dataset for the Locate task. As mentioned in Section~\ref{sec:dataset:set-lconvqa}, 
%% previous writing%% 
% To showcase the diverse applications of the models trained in Section~\ref{sec:set-consistency verification main section}, we use the Set-LConVQA dataset to evaluate the ability to detect inconsistent statements within a set (For simplicity, we call the name of this task as \textit{Locate} from now).  As mentioned in Section~\ref{sec:dataset:set-lconvqa}, 
%%%
% a key characteristic of Set-LConVQA is that in each inconsistent set $S_I$, there exists a specific two QA pairs, denoted as $s_i$ and $s_j$, such that the size-2 set $\{s_i, s_j\}$ is inconsistent. Moreover, 
through manual verification, we confirmed that in all sets in test data where the set size is at least four, there is exactly one element in $S_I$ whose removal renders the set consistent. For the Locate task, we evaluate and merge only those $S_C$ and $S_I$ sets that have a size of 4 or larger.

Similarly to the approach described in Section~\ref{sec:set-consistency verification main section}, for evaluation we use not only $S_{C}$ and $S_{I}$ but also sets constructed by merging two, three, or four different sets. This results in up to 14 possible dataset combinations with different labels (for example, $S_{C}$, $S_{I}$, $S_{CC}$, $\cdots$, $S_{IIII}$).

The Locate task leverages the model trained in Section~\ref{sec:set-consistency verification main section} without any additional task-specific training. For details on how the Locate task is performed using the LLM-based, binary classifier, and energy-based model architectures, please refer to Appendix~\ref{appendix: locate experiment setup detail}.

We measure performance using EM (Exact Match), Precision, Recall, and F1-score. For instance, in the case of $S_{IIII}$, an exact match is achieved only when all four inconsistent QA-pairs contained in the four inconsistent sets are correctly identified.
 Table \ref{tab:partb} shows the performance regarding the Locate task. Average and standard deviation is provided for each model (with different seeds) in table \ref{tab:parta}. As shown in Table \ref{tab:partb}, \scv~significantly outperforms LLM-based methods. Furthermore, \scv~also outperforms the set-level comparison \& binary classifier architecture model. This is a particularly impressive result given that Table \ref{tab:parta} shows only a marginal performance difference in set-consistency verification. \scv~learns fine-grained contrastive representations across multiple datasets, enabling it to capture the varying degrees of inconsistency within each set. This capability underlies its strong performance in accurately locating inconsistencies. 
% Refer to the appendix \ref{appendix: locate experiment result detail} for details on experimental results.

\section{Ablation Study}
\label{section: experiment additional results}

\subsection{Effect of Contrast Granularity}
\label{subsec:ablation study fine-grainedness contrast}

\begin{figure*}[t]
    \centering
    \includegraphics[width=1\linewidth]{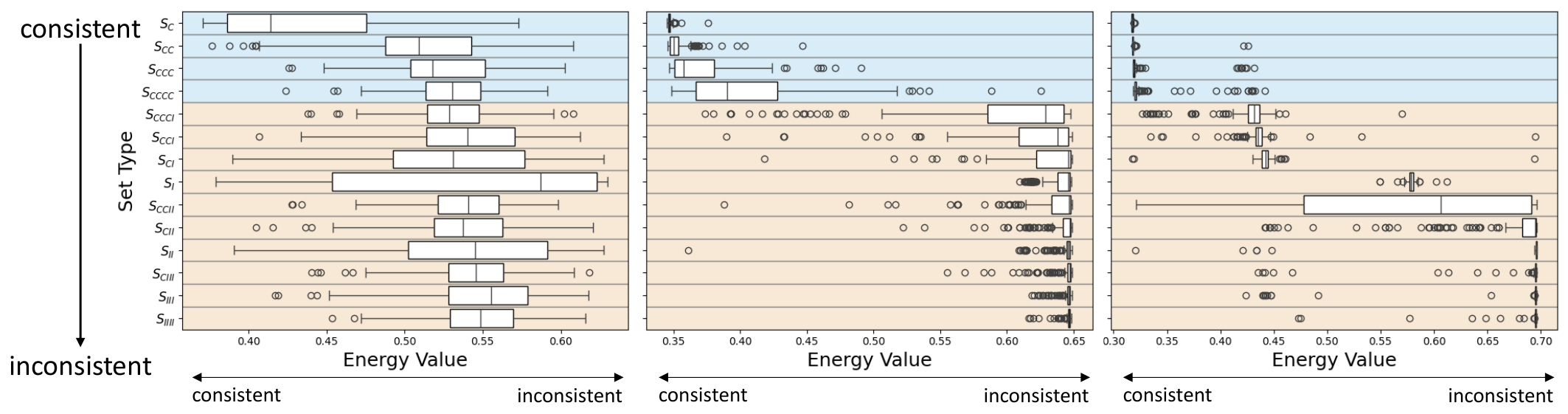}
    \caption{Box plots of energy values for various set types under different training regimes. \textbf{Left:} Training of \scv~ just contrasting $S_{C}$ vs. $S_{I}$. \textbf{Middle:} Training with six contrastive signals (contrasting all comparing sets with consistent vs. inconsistent labels), which does not capture the inconsistency degrees within inconsistent sets well. \textbf{Right:} Training with all eight contrastive signals, including the additional inconsistency degree contrasts, enables the model to distinguish sets more clearly—assigning higher energy to sets with a greater inclusion of inconsistent pairs (primary criterion) while moderating the energy for sets with a higher proportion of consistent pairs (secondary criterion). The set types are arranged from top to bottom in order of increasing inconsistency. Note that the more fine-grained contrastive signals are added, the energy levels more faithfully our intuition.}
    \label{fig:boxplots}
\end{figure*}

In Section~\ref{sec:fine-grained training \scv}, we introduced eight contrastive signals for training \scv. To evaluate their impact, we conducted an ablation study with three training regimes, as illustrated in Figure~\ref{fig:boxplots}:
\textbf{Left Panel:} The model is trained solely with the basic contrast $L_E(S_C, S_I)$. This limited contrast approach yields lower classifying performance.
\textbf{Middle Panel:} The model is trained with six contrasts—$L_E(S_C, S_I)$, $L_E(S_C, S_{CI})$, $L_E(S_C, S_{II})$, $L_E(S_{CC}, S_I)$, $L_E(S_{CC}, S_{CI})$, and $L_E(S_{CC}, S_{II})$—which compare sets with consistent labels against those with inconsistent labels, but do not capture variations within the inconsistent sets.
\textbf{Right Panel:} In addition to the six contrasts of middle panel, it also incorporates the two inconsistency degree contrasts $L_E(S_{CI}, S_I)$ and $L_E(S_I, S_{II})$. 
The eight contrasts training regime enables the model to capture the order of inconsistencies. Our criteria for ordering inconsistencies are threefold: the primary criterion, which deems sets containing more inconsistent pairs as more inconsistent; the secondary criterion, which moderates the overall inconsistency when consistent pairs dominate (i.e., the energy values remain intermediate rather than extreme); and the tertiary criterion, which asserts that even among consistent sets, the union of multiple consistent sets yields a more moderate consistency due to the inclusion of many neutral relationships, resulting in higher energy values compared to a pure $S_C$.

% Overall, the model trained with all eight contrasts (right panel) outperforms the others by better reflecting the nuanced degree of inconsistency across sets.

\subsection{Fine-Tuning Efficiency}
\label{sec:fine-tuning efficiency}
\begin{table}
\begin{center}
\scalebox{0.7}{\begin{tabular}{ccll}
\toprule
\multirow{3}{*}{\makecell[l]{Source}}& \multirow{3}{*}{\makecell[l]{Target (\#Data)}} & \multicolumn{2}{c}{Test Data} \\
  &  & \makecell[c]{Set-\\LConVQA} & \makecell[c]{Set-\\SNLI}\\
\midrule
\multirow{2}{*}{\makecell[c]{Set-\\LConVQA}} & Set-SNLI (100) & 0.938$_{\pm 0.025}$ & 0.666$_{\pm 0.008}$\\
 & Set-SNLI (200) & 0.942$_{\pm 0.056}$ & 0.727$_{\pm 0.007}$ \\
\cmidrule{2-4}
\multirow{2}{*}{\makecell[c]{Set-\\SNLI}} & Set-LConVQA (100) & 0.872$_{\pm 0.026}$ & 0.979$_{\pm 0.004}$ \\
 & Set-LConVQA (200) & 0.914$_{\pm 0.043}$ & 0.960$_{\pm 0.002}$ \\
\bottomrule
\end{tabular}}
\end{center}
\caption{Macro-F1 scores, with five random seed runs, for transferring the \scv~toward different domain. The table demonstrates the model's ability to retain performance on its original dataset while effectively generalizing to new domains with small amount of fine-tuning data.
}
\label{table:FineTuning_Binary_classification_simple}
\end{table}
Table \ref{table:FineTuning_Binary_classification_simple} shows the Macro-F1 scores of set-consistency verification task when fine-tuning the \scv~with a small amount of additional data from a different domain. Average and standard deviation for five different seed pairs are provided. The model retains strong performance on its original dataset while effectively adapting to new domains with minimal data.
Refer to the appendix \ref{appendix: fine-tuning efficiency} for experimental details.

\subsection{Verifying Inconsistencies in LLM Outputs}

Our \scv~can serve as a consistency evaluator that can be integrated with black-box LLMs to detect their inconsistent behaviors. While LLMs excel at various NLP tasks, they frequently produce outputs that are not logically coherent. For instance, an LLM might answer “yes” to “Does CO$_2$ increase the population of polar bears?' yet fail to provide a complementary response to “Does CO$_2$ decrease the population of polar bears?”.

To expose such inconsistencies, \cite{asai2020logic} proposed a data augmentation technique for the WIQA dataset \cite{tandon2019wiqa} that substitutes keywords with antonyms (e.g., “more” with “less”) to create question pairs with opposite expected answers. 
We evaluate the consistency of LLM on this augmented WIQA dataset to constrcut \textit{WIQA set consistency}, a new set-consistency dataset based on LLM answer generation.
% where question-answer pairs along with consistnecy label i. 
% we augmented the WIQA dataset and evaluated the consistency of LLM outputs on related question–answer pairs. 
In our construction, the LLM is queried independently for each related question to yield a set of question–prediction pairs whose logical coherence is then assessed in a rule-based manner. 
Our data collection with GPT-4o shows that it is 51.1\% consistent.

In order to validate the \scv's efficacy as a general consistency evaluator, we compare \scv's accuracy on WIQA set consistency data along with the Self-Check mechanism where LLM evaluates the consistency of its own response set.
As shown in Table~\ref{table:wiqa_external_module}, the consistency detection accuracy increases from 59.9\% with the self-check to 68.7\% when using \scv.
% As shown in Table~\ref{table:wiqa_external_module},  the Correct Rate—the proportion of cases where the LLM’s QA consistency aligns with the consistency detection outcome—increases from 59.9\% with the self-check to 68.7\% when using \scv. 
These results demonstrate that \scv~effectively identifies and flags inconsistent outputs, thereby functioning as a robust consistency evaluator when combined with existing LLMs.
Detailed descriptions of the evaluation methodology—including prompt design, the self-check process—are provided in appendix~\ref{appendix: external module evaluation details}.

\begin{table}[h]
\scalebox{0.75}{
\begin{tabular}{lcc}
\toprule
WIQA set consistency & Self-Check  & \scv~(Ours) \\
% & \makecell[c]{Overall} \\
\midrule
Consistent (51.1\%) & 79.6 &81.9\\
Inconsistent (48.9\%)& 39.2  & 54.8 \\
\midrule
Total Accuracy & \multicolumn{1}{c}{59.9} & \multicolumn{1}{c}{\textbf{68.7}} \\
\bottomrule
\end{tabular}}
\caption{ 
The table reports the accuracy of the set consistency verification (\%) on the WIQA set consistency dataset. 
% Both \scv~ and Self-Check show propensities toward predicting `consistent', however, \scv~ outperforms Self-Check in `inconsistency' label classification by a large margin.
% The table reports the conditional proportion of cases in which the LLM's own inference and its self-check mechanism (or \scv) agree on the consistency verification. For example, a score of 0.796 indicates that, among the 51.1\% of question–prediction pair sets where the LLM responded consistently, the self-check mechanism correctly verified consistency in 79.6\% of those cases. Overall Accuracy represents the percentage of all question–prediction pairs for which the self-check mechanism or \scv successfully verified consistency.
}
\label{table:wiqa_external_module}
\end{table}

\section{Conclusion}
\label{conclusion}

In this paper, we introduced the Set-Consistency Energy Network (\scv), a novel approach for verifying logical inconsistencies across multiple statements. 
% Unlike traditional methods that rely on pairwise comparisons, our model processes entire sets at once, capturing global dependencies that are often missed in conventional approaches.

Through extensive experiments on Set-LConVQA and Set-SNLI, we demonstrated that our energy-based framework significantly outperforms baseline methods, including LLM-based approaches and traditional classification models. The results show that our model excels in set-consistency verification and locating inconsistent statements within a set. Additionally, we explored its applicability as an consistency evaluator for verifying inconsistencies in LLM-generated outputs, showcasing its potential for real-world deployment.

% Our findings highlight the importance of set-level reasoning in consistency detection, and we believe that our approach can be extended to a broader range of NLP tasks that require logical coherence across multiple statements.

% By providing a scalable and effective method for detecting logical inconsistencies, \scv~contributes to the development of more reliable and interpretable AI systems, paving the way for enhanced reasoning capabilities in natural language processing.

% \section{Limitations}
% \label{sec:limitations}
\newpage
\section{Limitations and Potential Risks}

Our approach processes entire sets in a single pass, which, while enabling holistic consistency assessment, poses challenges when dealing with longer sets that may exceed the maximum token limit of current language models. 
Additionally, our two datasets Set-LConVQA, Set-SNLI are generated using a combination of rule-based methods and LLM-based transformations. Incorporating more diverse and representative data generation criteria could enable further advancements in this task. Moreover, evaluating the generalization capabilities of our approach across different domains remains an important direction for future research.

Other potential risks include a sensitivity to noise and edge cases present in real-world data. Addressing these issues will be critical for scaling the proposed method to broader applications.
\bibliography{custom}

\appendix
\onecolumn

\section{Details of \scv}
\label{appendix: training inference detail of energy network, threshold learning}
This section covers the topic of detail information on the training procedure using the eight contrasts, the conversion of a set $S$ into the input for \scv, and the selection of the threshold.

\subsection{Training Procedure Using the Eight Contrasts}
In Section \ref{sec:fine-grained training \scv}, we demonstrated that by appropriately merging $S_{C}$ and $S_{I}$, we can extend the set and train \scv~using eight distinct contrast methods. During training, the entire training dataset is treated as if it contains all eight types of contrast pairs (e.g., $(S_{C}, S_{I})$, $(S_{C}, S_{CI})$, $(S_{C}, S_{II})$, etc.). Whenever a sample corresponding to a specific contrast type is drawn from the dataset, its loss $L_{E}(\cdot, \cdot)$ is computed based on that contrast.

\subsection{Conversion of $S$ into the Input for \scv}
\label{appendix: conversion of set into the input for energy network}
For a set $S = \{s_{1}, s_{2}, \cdots, s_{n}\}$ composed of individual sentences, the input to \scv~is formed by concatenating all sentences $s_{i}$ and prepending a CLS token at the beginning. For example, consider a set $S$ from the Set-SNLI dataset defined as:
\begin{align}
\nonumber
S &= \{\text{``Either the train arrives at 8 AM or it arrives at 9 AM.''},\\ 
\nonumber
&\quad \text{``The train does not arrive at 8 AM.''}, \\
\nonumber
&\quad \text{``The train does not arrive at 9 AM.''}\}
\end{align}
In the case of the RoBERTa-base model, the CLS token is represented as \texttt{<s>}, so the input for set $S$ is:
\begin{align}
\nonumber
&\texttt{<s> Either the train arrives at 8 AM or it arrives at 9 AM.}\\
\nonumber
&\texttt{The train does not arrive at 8 AM. The train does not arrive at 9 AM.}    
\end{align}

In contrast, for a set $S = \{(x_{i}, y_{i})\}_{i=1}^{n}$ composed of QA pairs, we concatenate each $x_{i}$ and $y_{i}$ using the delimiter \textit{``The answer is''}. After concatenating all the QA pairs, we prepend a CLS token at the beginning. For example, consider a set $S$ from the Set-LConvqa dataset defined as:
\begin{align}
\nonumber
S = \{\text{(``what color is desk?'', ``brown'')}, \text{(``is desk brown?'', ``yes'')}\}
\end{align}
With RoBERTa-base, where the CLS token is \texttt{<s>}, the input for set $S$ becomes:
\begin{align}
\nonumber
\texttt{<s> what color is desk? The answer is brown. is desk brown? The answer is yes.}    
\end{align}

When converting a set $S$ into the input for \scv, all elements of $S$ are shuffled before being fed into the model.

\subsection{Selection of Threshold}
\label{appendix: selection of Threshold, details of \scv}
The dataset is split into training data, validation1 data, validation2 data, and test data. At the end of each training epoch, the threshold is updated using validation1 data. A set is classified as consistent if its energy value is below the threshold; otherwise, it is classified as inconsistent. The threshold is determined by maximizing macro classification accuracy across validation1 data consisting of up to two combined sets: $S_C$, and  $S_{CC}$ as consistent sets, while $S_I$, $S_{CI}$, and $S_{II}$ are labeled as inconsistent. Finally, validation2 data is used for hyperparameter tuning.

\section{Set-LConVQA Dataset}
\label{appendix: set lconvqa details}
The Set-LConVQA dataset is derived from the original LConVQA \cite{ray-etal-2019-sunny} dataset which is publicly available\footnote{https://arijitray1993.github.io/ConVQA/}, where for each image, both consistent and inconsistent sets of QA pairs are generated. For example, a consistent set is represented as follows:

\begin{verbatim}
"consistent": [
    [
        {"question": "what color is desk?", "answer": "brown"},
        {"question": "is desk brown?", "answer": "yes"},
        {"question": "is desk pink?", "answer": "no"}
    ]
]
\end{verbatim}

In contrast, an inconsistent set is always composed of exactly two QA pairs, as in this example:

\begin{verbatim}
"inconsistent": [
    [
        {"question": "what color is desk?", "answer": "brown"},
        {"question": "is desk brown?", "answer": "no"}
    ],
    [
        {"question": "what color is desk?", "answer": "brown"},
        {"question": "is desk pink?", "answer": "yes"}
    ]
]
\end{verbatim}

Our rule-based validation of the LConVQA dataset revealed two key observations:
\begin{enumerate}
    \item For every consistent set derived from a given image, there exists at least one inconsistent set such that all questions that appear in that inconsistent set are also present in the corresponding consistent set.
    \item In the inconsistent sets, one answer differs from the answer in the consistent set for the corresponding question.
\end{enumerate}

In other words, an inconsistent set can be constructed from a consistent set by altering the answer of a single QA pair while keeping all the questions unchanged. Formally, given a consistent set 
\[
S_{C} = \{(q_1, a_1), (q_2, a_2), \dots, (q_n, a_n)\},
\]
an inconsistent set can be obtained as 
\[
S_{I} = \{(q_1, a_1), \dots, (q_i, a_i^*), \dots, (q_n, a_n)\},
\]
where \(a_i^*\) differs from \(a_i\). Notably, the generated inconsistent set \(S_{I}\) encompasses the inconsistent sets originally present in the LConVQA dataset. As defined in Section~\ref{sec:definitions}, a set \(S\) is deemed \textbf{inconsistent} if there exists a subset \(\tilde{S} \subseteq S\) with \(|\tilde{S}| \ge 2\) such that the statements in \(\tilde{S}\) are logically contradictory. Hence, by our definition, \(S_{I}\) qualifies as an inconsistent set.
On average, each set contains 3.43 QA pairs, with a maximum of six.

Because the element-wise verification strategy described in Section~\ref{sec:Experiments} is trained on sets of size 2, we can construct a dedicated element-wise training dataset from the generated Set-LConVQA dataset. Specifically, the inconsistent sets \(S_{I}\) in the original LConVQA dataset (which are of size 2) can be directly used for element-wise verification. Similarly, since every inconsistent set derived from a consistent set \(S_{C}\) involves altering only the answer while keeping all questions identical, the corresponding QA pairs can be extracted to form the element-wise consistent set for training with the element-wise strategy.

\section{Set-SNLI dataset}
\label{appendix:set-snli_dataset}

\subsection{Dataset Creation}
Inspired by previous work \cite{nakamura-etal-2023-logicattack}, we construct the Set-SNLI dataset by transforming existing pairwise SNLI datasets\footnote{SNLI dataset is publicly available, https://nlp.stanford.edu/projects/snli/. This dataset takes Creative Commons Attribution-ShareAlike 4.0 International License, allowing remix, transform, and build upon the material for any purpose.} according to predefined rules. The generation process consists of the following steps:

\textbf{1. Generating Negations}: Using a large language model (LLM), we generate negated versions of the premises and hypotheses in the original dataset. The prompt used for negation generation is included in the appendix \ref{appendix:set nli negation prompt}.

\textbf{2. Constructing Sentence Sets}: One or two pairs of premises and hypotheses, along with their negations, are combined according to predefined rules. 

The rules applied vary based on the label of the original premise-hypothesis relationship and the number of seed pairs used. Detailed rules are outlined in the appendix \ref{appendix:set nli rules}. 
%더 나은 서술방식?
%Logic attack에 대한 언급?
Some rules employ first-order logic inference to guarantee entailment between sentences, while others guarantee contradiction. Additional techniques involve leveraging sentence equivalence or combining previously generated sets to create larger sentence groups.

%transition?
Sentences connected with conjunctions (AND) are always split so that each element in a set remains an atomic sentence. The definition of "atomic" may vary, but structures involving disjunctions (OR) or conditionals (if-then) are allowed. However, AND operations are disallowed to prevent arbitrary manipulation of set sizes by linking sentences with AND. 
% 이유를 더 잘 설명하기 
Although the above dataset creation process can be applied to any single sentence based conventional NLI datasets, for our implementation, we derive our dataset from the test set of SNLI dataset.

\subsection{Dataset Structure}
Each data instance consists of the following: a set of two or more sentences, a label, i.e., either consistent or inconsistent, difficulty annotation, and set size. 
The difficulty annotation is categorized as either "medium" or "easy". The "easy" category is assigned only to contradictory sets that include a sentence and its direct negation, which we consider trivial to identify as contradictory.
On average, each set contains 3.48 sentences, with a maximum of five. 
% Sentence length varies depending on the original dataset used for transformation. For example, when derived from the SNLI test set, the average sentence length is 17.22 words, approximately 1.6 times longer than the original dataset (10.72 words). 

\subsection{Negation generation prompt}
\label{appendix:set nli negation prompt}
In order to create negated versions of premises and hypotheses, we instruct an LLM (in our case, gpt4o-mini) with the prompt provided in the Table \ref{tab:set_nli_negation_prompt}

\begin{table*}[h]
\centering
% \scriptsize
\scalebox{0.6}{
\begin{tabular}{@{}l@{}}
\toprule
Instruction \\ \midrule
\begin{tabular}[c]{@{}l@{}}Premise: An older man wearing a salon drape getting a haircut.\\ Hypothesis: A man gets a haircut.\\ Negation of Hypothesis: No man gets a haircut.\\ Negation of Premise: No older man wearing a salon drape is getting a haircut.\\ Premise: A man with glasses sitting at a restaurant staring at something that is not shown.\\ Hypothesis: A man at a restaurant.\\ Negation of Hypothesis: No man at a restaurant.\\ Negation of Premise: No man with glasses sitting at a restaurant staring at something that is not shown.\\ Premise: The school is having a special event in order to show the American culture on how other cultures are dealt with in parties.\\ Hypothesis: A school is hosting an event.\\ Negation of Hypothesis: No school is not hosting an event.\\ Negation of Premise: The school is not having a special event to show the American culture or how other cultures are dealt with in parties.\\ Premise: High fashion ladies wait outside a tram beside a crowd of people in the city.\\ Hypothesis: Women are waiting by a tram.\\ Negation of Hypothesis: No women are waiting by a tram.\\ Negation of Premise: No high fashion ladies wait outside a tram beside a crowd of people in the city.\\ Premise: People waiting to get on a train or just getting off.\\ Hypothesis: There are people waiting on a train.\\ Negation of Hypothesis: There are no people waiting on a train.\\ Negation of Premise: No people are waiting to get on a train or just getting off.\\ Premise: A couple play in the tide with their young son.\\ Hypothesis: The family is outside.\\ Negation of Hypothesis: The family is not outside.\\ Negation of Premise: A couple does not play in the tide with their young son.\\ $\{$ More examples omitted $\}$ \\ \\ Take a closer look at these tricky cases; 'A' or 'An' needs to change to 'No' even if the sentence does not start with the subject, and also when two clauses exist in a single sentence.\\ Premise: Under a blue sky with white clouds, a child reaches up to touch the propeller of a plane standing parked on a field of grass.\\ Hypothesis: A child is reaching to touch the propeller of a plane.\\ Negation of Hypothesis: Under a blue sky with white clouds, no child reaches up to touch the propeller of a plane standing parked on a field of grass.\\ Negation of Premise: No child is reaching to touch the propeller of a plane\\ Premise: A man in a green shirt is singing karaoke while a young woman with long brownish hair stands by and listens.\\ Hypothesis: A man is singing a song.\\ Negation of Hypothesis: No man is singing a song.\\ Negation of Premise: No man in a green shirt is singing karaoke while no young woman with long browish hair stands by and listens.\\ \\ When a sentence starts with subjects specified with 'the' or 'this', verb should be negated rather than the subject.\\ Premise: A woman with a green headscarf blue shirt and a very big grin.\\ Hypothesis: The woman is very happy.\\ Negation of Hypothesis: The woman is not very happy.\\ Negation of Premise: No woman with a green headscarf blue shirt and a very big grin.\\ Premise: The man walks among the large trees.\\ Hypothesis: The man walks among trees.\\ Negation of Hypothesis: The man does not walk among trees.\\ Negation of Premise: The man does not walk among the large trees.\\ \\ Please note that given a compound sentence connecting two or more clauses with an 'and' or a comma, each of the clauses need to be negated, connected by an 'or'.\\ Premise: Four people are near a body of water, two people walk on a sidewalk.\\ Hypothesis: There are people outdoors.\\ Negation of Hypothesis: There are no people outdoors\\ Negation of Premise: No four people are near a body of water or no two people walk on a sidewalk.\\ \\ Construct Negation of Hypothesis and Negation of Premise by following the examples above.\\ Given the condtion that Premise entails Hypothesis, Negation of Premise needs to entail Negation of Hypothesis.\end{tabular} \\ \bottomrule
\end{tabular}}
\caption{Instruction used to create negation of premise and hypothesis of the original NLI dataset. The few shot examples in the first part of the instruction are taken from \citealp{nakamura-etal-2023-logicattack}.}
\label{tab:set_nli_negation_prompt}
\end{table*}

\subsection{Dataset creation rules}
\label{appendix:set nli rules}
In this section, we provide rules used to create a set-based NLI dataset from a conventional pair-wise SNLI dataset. For convenience, we divide the section in terms of the number and label of seed pair(s)(in other words, the original premise-hypothesis pair in SNLI dataset) we use to create each set. 
For tables \ref{tab:single_entail}, \ref{tab:single_contradiction}, \ref{tab:single_neutral}, \ref{tab:double_entail}, the singleton rules are motivated by \cite{nakamura-etal-2023-logicattack}, and we generated additional rules from them (represented by the union mark $\cup$).
% For instance, if it's a single seed pair case, then we take one pair of (premise, hypothesis) from a conventional dataset and modify the premise and hypothesis to create a set of statements with either "consistent" or "inconsistent" label. If it's a double seed pair case, then we take two initial pairs, modify the premise and hypothesis in them, and combine them to create a set of statements with either "consistent" or "inconsistent" label. 

\subsubsection{Rules for deriving sets from a single entailment seed pair}
Please refer to Table \ref{tab:single_entail}.
\subsubsection{Rules for deriving sets from a single contradiction seed pair}
Please refer to Table \ref{tab:single_contradiction}.
\subsubsection{Rules for deriving sets from a single neutral seed pair}
Please refer to Table \ref{tab:single_neutral}.
\subsubsection{Rules for deriving sets from two entailment seed pairs}
Please refer to Table \ref{tab:double_entail}.

\subsubsection{Rules for Deriving Sets for Element-Wise Verification Strategy}
\label{appendix: set-snli elementwise training dataset construction}

Since the element-wise verification strategy (discussed in Section~\ref{sec:Experiments}) is trained on sets of size 2, we  construct a dedicated element-wise dataset using the generated Set-SNLI dataset.

Within a given set, a pair of propositions that exhibit a logically inconsistent relationship can be selected when the single entailment pair (as described in Table~\ref{tab:single_entail}) satisfies one of the following conditions between $p$ and $h$:
\begin{enumerate}
    \item $p$, $\neg p$
    \item $h$, $\neg h$
    \item $p$, $\neg h$
    \item $p \vee h$, $\neg h$
\end{enumerate}

For each single entailment seed pair $(p, h)$, an inconsistent set $S_{I}$ can be created by applying the above rules. Conversely, for constructing a consistent set $S_{C}$, any size-2 subset extracted from a consistent set qualifies as a consistent set, according to the definition provided in Section~\ref{sec:definitions}.

\begin{table*}
\centering
\tiny
\begin{tabular}{|c|l|l|c|c|c|}
\hline
No. & Rule & Description & Set Size & Label & Difficulty \\
\hline
1& $\{p_1 \rightarrow h_1,\neg h_1 \rightarrow \neg p_1\}$&Transportation& 2& Consistent& Medium \\
2& $\{p_1 \rightarrow h_1,\neg p_1 \lor h_1\}$&Material Implication& 2& Consistent& Medium \\
3& $\{p_1 \rightarrow h_1, h_1\}$&Split Hypothesis of Rule 2 (1)& 2& Consistent& Medium \\
4& $\{p_1 \rightarrow h_1, \neg p_1\}$&Split Hypothesis of Rule 2 (2)& 2& Consistent& Medium \\
5& $\{p_1 \rightarrow h_1, p_1,h_1\}$&Modus Ponens& 3& Consistent& Medium \\
6& $\{p_1 \rightarrow h_1,\neg h_1, \neg p_1\}$&Modus Tollens& 3& Consistent& Medium \\
7& $\{p_1 \lor h_1,\neg h_1, p_1\}$&Disjunctive Syllogism (1)& 3& Consistent& Medium \\
8& $\{p_1 \lor h_1,\neg p_1, h_1\}$&Disjunctive Syllogism (2)& 3& Consistent& Medium \\
9& $\{p_1 \rightarrow h_1,\neg h_1 \rightarrow \neg p_1, \neg p_1 \lor h_1\}$& Rule 1 $\cup$ 2 & 3& Consistent& Medium \\
10& $\{p_1 \rightarrow h_1,\neg h_1 \rightarrow \neg p_1, h_1\}$& Rule 1 $\cup$ 3& 3& Consistent& Medium \\
11& $\{p_1 \rightarrow h_1,\neg h_1 \rightarrow \neg p_1, \neg p_1\}$&Rule 1 $\cup$ 4 & 3& Consistent& Medium \\
12& $\{p_1 \rightarrow h_1,\neg p_1 \lor h_1, h_1\}$&Rule 2 $\cup$ 3 & 3& Consistent& Medium \\
13& $\{p_1 \rightarrow h_1,\neg p_1 \lor h_1, \neg p_1\}$&Rule 2 $\cup$ 4 & 3& Consistent& Medium \\
14& $\{p_1 \rightarrow h_1,\neg p_1, h_1\}$&Rule 3 $\cup$ 4& 3& Consistent& Medium \\
15& $\{p_1 \rightarrow h_1, \neg h_1 \rightarrow \neg p_1, p_1, h_1\}$&Rule 1 $\cup$ 5& 4& Consistent& Medium \\
16& $\{p_1 \rightarrow h_1, \neg h_1 \rightarrow \neg p_1, \neg p_1, \neg h_1\}$& Rule 1 $\cup$ 6& 4& Consistent& Medium \\
17& $\{p_1 \rightarrow h_1,\neg p_1 \lor h_1, p_1, h_1\}$&Rule 2 $\cup$ 5& 4& Consistent& Medium \\
18& $\{p_1 \rightarrow h_1,\neg p_1 \lor h_1, \neg p_1, \neg h_1\}$&Rule 2 $\cup$ 6& 4& Consistent& Medium \\
19& $\{p_1 \rightarrow h_1, \neg p_1 \lor h_1, \neg h_1 \rightarrow \neg p_1, h_1\}$&Rule 1 $\cup$ 2 $\cup$ 3& 4& Consistent& Medium \\
20& $\{p_1 \rightarrow h_1, \neg p_1 \lor h_1, \neg h_1 \rightarrow \neg p_1, \neg p_1\}$&Rule 1 $\cup$  2 $\cup$ 4& 4& Consistent& Medium \\
21& $\{p_1 \rightarrow h_1, \neg h_1 \rightarrow \neg p_1, \neg p_1,  h_1\}$&Rule 1 $\cup$ 3 $\cup$ 4& 4& Consistent& Medium \\
22& $\{p_1 \rightarrow h_1, \neg p_1 \lor h_1, \neg p_1, h_1\}$&Rule 2 $\cup$ 3 $\cup$ 4& 4& Consistent& Medium \\
23& $\{p_1 \rightarrow h_1, \neg p_1 \lor h_1, \neg h_1 \rightarrow \neg p_1, p_1, h_1\}$&Rule 1 $\cup$  2 $\cup$ 5& 5& Consistent& Medium \\
24& $\{p_1 \rightarrow h_1, \neg p_1 \lor h_1, \neg h_1 \rightarrow \neg p_1, \neg p_1, \neg h_1\}$&Rule 1 $\cup$ 2 $\cup$ 6& 5& Consistent& Medium \\
25& $\{p_1 \rightarrow h_1, \neg p_1 \lor h_1, \neg h_1 \rightarrow \neg p_1, \neg p_1, h_1\}$&Rule 1 $\cup$ 2 $\cup$ 3 $\cup$ 4& 5& Consistent& Medium \\
\hline
26& $\{p_1, \neg h_1\}$& Negate Hypothesis (1) & 2& Inconsistent& Medium \\
27& $\{p_1, \neg h_1, p_1 \rightarrow h_1\}$& Negate Hypothesis (2) & 3& Inconsistent& Medium \\
28& $\{p_1 \lor h_1, \neg p_1, \neg h_1\}$&Negate Hypothesis of Rule 7& 3& Inconsistent& Medium \\
29& $\{p_1 \lor h_1, p_1 \rightarrow h_1, \neg h_1\}$&Implicit Negate Hypothesis of Rule 7& 3& Inconsistent& Medium \\
30& $\{p_1 \rightarrow h_1, \neg h_1, \neg p_1, p_1\}$&Rule 6 $\cup$ Rule 26 & 4& Inconsistent& Easy \\
31& $\{p_1 \rightarrow h_1, \neg h_1, \neg p_1, h_1\}$&Rule 6 $\cup$ Rule 3& 4& Inconsistent& Easy \\
32& $\{p_1 \rightarrow h_1, p_1, h_1, \neg p_1\}$&Rule 5 $\cup$ Rule 4& 4& Inconsistent& Easy \\
33& $\{p_1 \rightarrow h_1, p_1, h_1, \neg h_1\}$&Rule 5 $\cup$ Rule 26& 4& Inconsistent& Easy \\
34& $\{p_1 \rightarrow h_1, \neg h_1 \rightarrow \neg p_1, p_1, \neg h_1\}$& Rule 1 $\cup$ Rule 26  & 4& Inconsistent& Medium \\
35& $\{p_1 \rightarrow h_1, \neg p_1 \lor h_1, p_1, \neg h_1\}$& Rule 2 $\cup$ Rule 26 & 4& Inconsistent& Medium \\
36& $\{p_1 \rightarrow h_1, \neg h_1, \neg p_1, p_1, h_1\}$& Rule 6 $\cup$ Rule 5 & 5& Inconsistent& Easy \\
\hline
    \end{tabular}
    \caption{List of rules used to create sets from a single entailment seed pair. The seed pair is comprised of a premise and a hypothesis, i.e., $\{p_1,h_1\}$, that are in entailment relationship $p_1\rightarrow h_2$. }
    \label{tab:single_entail}
\end{table*}

\begin{table*}
\centering
\tiny
\begin{tabular}{|c|l|l|c|c|c|}
\hline
No. & Rule & Description & Set Size & Label & Difficulty \\
\hline
1& $\{\neg p_1, h_1\}$&Negate Premise & 2& Consistent& Medium \\
2& $\{p_1, \neg h_1\}$&Negate Hypothesis & 2& Consistent& Medium \\
3& $\{p_1 \lor h_1,\neg h_1, p_1\}$&Disjunctive Syllogism (1)& 3& Consistent& Medium \\
4& $\{p_1 \lor h_1,\neg p_1, h_1\}$&Disjunctive Syllogism (2)& 3& Consistent& Medium \\
\hline
5& $\{p_1 \lor h_1, p_1, h_1\}$& Disjunction $\cup$ Seed Pair & 3& Inconsistent& Medium \\
6& $\{p_1 \lor h_1, \neg p_1, \neg h_1\}$& Negate Hypothesis of Rule 3& 3& Inconsistent& Medium \\
\hline
    \end{tabular}
    \caption{List of rules used to create sets from a single contradiction seed pair. The seed pair consists of a premise and a hypothesis, i.e. $\{p_1,h_1\}$, that are in a contradiction relationship $p_1\land h_2\rightarrow \bot$.}
    \label{tab:single_contradiction}
\end{table*}

\begin{table*}
\centering
\tiny
\begin{tabular}{|c|l|l|c|c|c|}
\hline
No. & Rule & Description & Set Size & Label & Difficulty \\
\hline
1& $\{p_1 \lor h_1,\neg h_1, p_1\}$&Disjunctive Syllogism 1& 3& Consistent& Medium \\
2& $\{p_1 \lor h_1,\neg p_1, h_1\}$&Disjunctive Syllogism 2& 3& Consistent& Medium \\
\hline
3& $\{p_1 \lor h_1, \neg p_1, \neg h_1\}$& Negate Hypothesis of Rule 1 & 3& Inconsistent& Medium \\
\hline
    \end{tabular}
    \caption{List of rules used to create sets from a single neutral seed pair. The seed pair consists of a premise and a hypothesis, i.e. $\{p_1,h_1\}$, that are in neutral relationship, i.e., $|p_1\land h_1|>0$ and $p_1\not\to h_1$ and $h_1\not\to p_1$.}
    \label{tab:single_neutral}
\end{table*}

\begin{table*}
\centering
\tiny
\begin{tabular}{|c|l|l|c|c|c|}
\hline
No. & Rule & Description & Set Size & Label & Difficulty \\
\hline
1& $\{p_1 \rightarrow h_1, p_2 \rightarrow h_2, p_1 \lor p_2, h_1 \lor h_2\}$& Constructive Dilemma & 4& Consistent& Medium \\
2& $\{p_1 \rightarrow h_1, p_2 \rightarrow h_2, \neg h_1 \lor \neg h_2, \neg p_1 \lor \neg p_2\}$& Destructive Dilemma & 4& Consistent& Medium \\
3& $\{p_1 \rightarrow h_1, p_2 \rightarrow h_2, p_1 \lor \neg h_2, h_1 \lor \neg p_2\}$& Bidirectional Dilemma & 4& Consistent& Medium \\
\hline
4& $\{p_1 \rightarrow h_1, p_2 \rightarrow h_2, p_1 \lor p_2, \neg h_1, \neg h_2\}$& Negate Hypothesis of Constructive Dilemma & 5& Inconsistent& Medium \\
5& $\{p_1 \rightarrow h_1, p_2 \rightarrow h_2, \neg h_1 \lor \neg h_2, p_1, p_2\}$& Negate Hypothesis of Destructive Dilemma & 5& Inconsistent& Medium \\
6& $\{p_1 \rightarrow h_1, p_2 \rightarrow h_2, p_1 \lor \neg h_2, \neg h_1,  p_2\}$& Negate Hypothesis of Bidirectional Dilemma & 5& Inconsistent& Medium \\

\hline
    \end{tabular}
    \caption{List of rules used to create sets from two entailment seed pairs, $\{p_1,h_1\}$ and $\{p_2,h_2\}$. Each seed pair is in an entailment relationship, i.e. $p_1\rightarrow h_1$ and $p_2\rightarrow h_2$.}
    \label{tab:double_entail}
\end{table*}

\begin{table*}[h]
\centering
\begin{tabular}{@{}l@{}}
\toprule
Original Premise \& Hypothesis \\ \midrule
\begin{tabular}[c]{@{}l@{}}$p$: A couple walk hand in hand down a street.\\ $h$: A couple is walking together.\\ Original label: entailment\end{tabular} \\ \midrule\midrule
Resulting Sets \\ \midrule
\begin{tabular}[c]{@{}l@{}}⇒ Applied rule: $\{p \rightarrow h,\neg h, \neg p\}$ (Modus Tollens)\\ ⇒ Set: \{"If a couple walk hand in hand down a street, then a couple is   walking together.",\\      "No couple is walking together.",\\      "No couple walks hand in hand down a street."\}\\ ⇒ Label: consistent\end{tabular} \\ \midrule
\begin{tabular}[c]{@{}l@{}}⇒ Applied rule: $\{p_1 \lor h_1, p_1 \rightarrow h_1, \neg h_1\}$ (Implicit   Negate Hypothesis of Disjunctive Syllogism)\\ ⇒ Set: \{"If a couple walk hand in hand down a street, then a couple is   walking together.",\\      "No couple is walking together.",\\      "Either a couple walk hand in hand down a street, or a couple is   walking together."\}\\ ⇒ Label: inconsistent\end{tabular} \\ \bottomrule
\end{tabular}
\caption{Examples of generating Set-NLI data from a conventional SNLI dataset. The original premises and hypotheses are sampled from the SNLI dataset. Note that Set-SNLI data is derived not only from premise-hypothesis pairs with an entailment relationship but also from those labeled as contradiction or neutral. We create both consistent and inconsistent sets from the seed pairs.}
\label{tab:set-snli-examples}
\end{table*}

\section{Experiment}

\subsection{Set-Consistency Verification}
\label{appendix: set-consistency verification}
In this section, we present further detailed experimental results related to set-consistency verification.

\subsubsection{Experiment Setup Detail}
\label{appendix: set-consistency verification experiment setting detail}

In this section, we discuss the experimental setups employed for each model architecture and verification strategy.

\paragraph{LLM Model Architecture}
For the LLM-based model architecture, we employ chain-of-thought (CoT) based prompting to request a consistency classification for a given set. In the \emph{set-level verification strategy}, the prompt is formulated as follows:

\textbf{Prompt:}
\begin{quote}
Tell me whether the following question-answer pairs are consistent or inconsistent.

\{Few-shot CoT examples (5-shot)\}

\{Problem\}

Please think step by step: first, clearly articulate your thought process; then, provide your final consistency judgment by choosing either `consistent' or `inconsistent' after the `Consistency:' mark.
\end{quote}

% A corresponding figure (e.g., Figure~\ref{fig:LLM_SetLevel}) illustrates this experimental setup.

For the \emph{element-wise verification strategy}, the form of prompt and the number of few-shot CoT examples remain identical. However, there are two key differences:
\begin{enumerate}
    \item When evaluating a set of size $N$, every possible pair of elements (i.e., $\frac{N(N-1)}{2}$ pairs) is compared for consistency. If even one pair is determined to be inconsistent, the entire set is classified as inconsistent.
    \item The few-shot CoT examples are constructed exclusively from sets consisting of two elements.
\end{enumerate}

\paragraph{Binary Classification Model Architecture}  

In the \textit{set-level verification strategy}, the model's input is the entire set (for details on how to convert a set into the model's input, please refer to Appendix~\ref{appendix: conversion of set into the input for energy network}; although this model is not \scv, the method for converting the set into the input is the same). The model outputs a 2-dimensional vector, where each component represents the score for either "consistent" or "inconsistent." 

In the \textit{element-wise verification strategy}, the process is similar; however, when evaluating a set of size $N$, every possible pair of elements (i.e., $\frac{N(N-1)}{2}$ pairs) is compared for consistency. This amounts to repeatedly assessing sets of size 2. If even one pair is determined to be inconsistent, the entire set is classified as inconsistent.

Typically, when making predictions from a 2-dimensional vector, it is common to select the class corresponding to the higher score. However, in our experiment setting, a threshold is also be learned for the binary classification model. The output scores for "consistent" and "inconsistent" are passed through a softmax function, and a threshold is learned on the "inconsistent" class score in the same manner as described in Appendix~\ref{appendix: selection of Threshold, details of \scv}. This learned threshold is then used during inference.

For the binary classification model architecture, training is conducted using a cross-entropy loss function. Under the set-level verification strategy, we treat $S_{C}$ and $S_{CC}$ as consistent label data and $S_{I}$, $S_{CI}$, and $S_{II}$ as inconsistent label data. For the element-wise verification strategy, we used the training data described in Appendices~\ref{appendix: set lconvqa details} and \ref{appendix:set-snli_dataset}.

\paragraph{Energy-Based Model Architecture}  
For the energy-based model architecture using the \textit{set-level verification strategy}, the model is implemented as \scv. Details regarding the training and inference procedures, as well as threshold selection, can be found in Appendix~\ref{appendix: training inference detail of energy network, threshold learning}. In the \textit{element-wise verification strategy}, the procedure is analogous to that of the binary classification model architecture: for a set of size $N$, pairs of elements are extracted to form sets of size 2, and the consistency of each pair is evaluated. If even one pair is determined to be inconsistent, the entire set is classified as inconsistent.

We used Adam optimizer \cite{diederik2014adam}, RTX A6000 GPU\footnote{https://www.nvidia.com/content/dam/en-zz/Solutions/design-visualization/quadro-product-literature/proviz-print-nvidia-rtx-a6000-datasheet-us-nvidia-1454980-r9-web\%20(1).pdf}, and RoBETa-base \cite{liu2019roberta} model (125M) for all of trainings (for both binary classification and energy-based model architectures). We used grid search to find learning rate, one oe $1$, $1e-1$, $\cdots$, $1e-7$. Learning rate of $1e-6$ is used for the loss function of binary classifier models, and learning rate of $1e-5$ is used for the loss function of energy-based models. The margin of $0.01$ is used in $L_{E}(S_{C}, S_{I}) = [E_{\theta}(S_{C}) - E_{\theta}(S_{I}) + \alpha]_{+}$ as $\alpha$ for energy-based models.

For the energy-based model architecture, training is conducted using a margin-based loss function. The training procedure of set-level verification strategy is described in \ref{sec:fine-grained training \scv}. For the element-wise verification strategy, we used the training data described in Appendices~\ref{appendix: set lconvqa details} and \ref{appendix:set-snli_dataset}.

\subsubsection{Maximum Tolerance Rate for Element-wise Verification strategy}
\label{appendix: section maximum tolerance rate}
In a pairwise comparison approach, if any one-to-one pair comparison detects an inconsistency, the entire document is theoretically classified as inconsistent. This raises the question: to what extent should a certain level of inconsistency be tolerated (even if this is not the ideal approach) to achieve optimal performance in set-consistency verification?

To investigate this, we define the \emph{Maximum Tolerance Rate} (MTR). For a set $S$ with $|S| = N$, there are $\frac{N(N-1)}{2}$ possible pairwise comparisons. The MTR, denoted by $p$, represents the maximum allowable proportion of inconsistent pairs such that the set $S$ is still classified as consistent. Formally, if
\[
\frac{\text{number of predicted one-to-one pairs as inconsistent}}{\text{total number of pairs }(=\frac{N(N-1)}{2})} \leq p,
\]
then $S$ is deemed consistent; otherwise, it is classified as inconsistent. Note that when MTR $= 0$, it is theoretically the most appropriate decision rule (i.e., if even one one-to-one pair is predicted to be inconsistent, the entire set $S$ is classified as inconsistent). As the value of MTR increases, the likelihood of classifying set $S$ as consistent also increases.

\begin{figure}[H]
    \centerline{\includegraphics[width=\columnwidth]{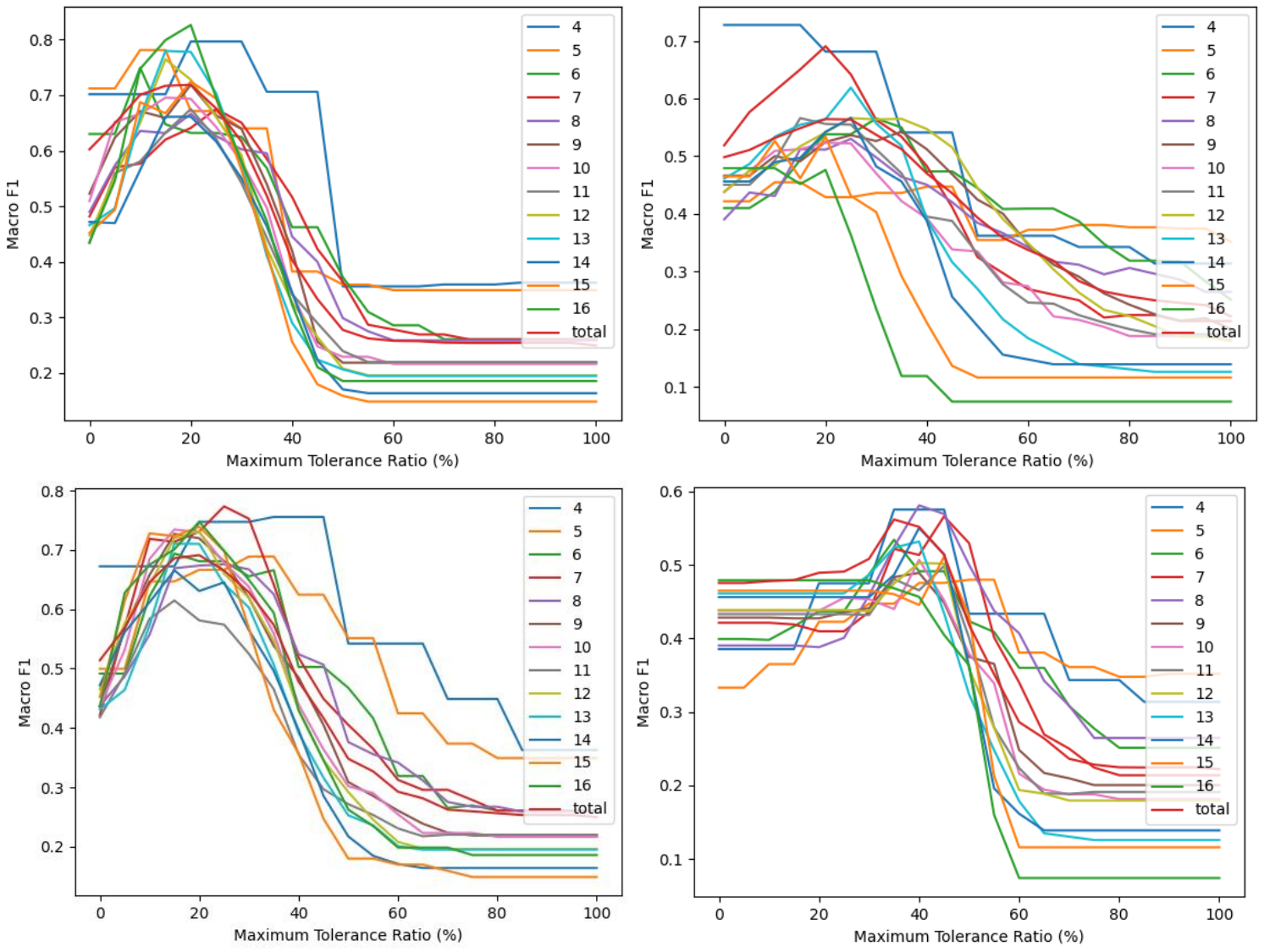}}
    \caption{Figure shows the Macro-F1 score of set-consistency verification task  across various MTR values and set sizes. \textbf{Left} figures represents the Macro-F1 score for Set-LConVQA dataset, and \textbf{Right} figures represents the Macro-F1 score for Set-SNLI dataset. Also, figures on the \textbf{Top} represents the Macro-F1 score for binary classifier models, and figures on the \textbf{Bottom} represents the Macro-F1 score for energy-based models.}
    \label{fig:lconvqa_f1} 
\end{figure}

Figure \ref{fig:lconvqa_f1} shows the Macro-F1 score for the  element-wise verification strategies across various MTR values and set sizes. \textbf{Left} figures represents the Macro-F1 score for Set-LConVQA dataset, and \textbf{Right} figures represents the Macro-F1 score for Set-SNLI dataset. Also, figures on the \textbf{Top} represents the Macro-F1 score for binary classifier models, and figures on the \textbf{Bottom} represents the Macro-F1 score for energy-based models. As the figure illustrates, there exists a point—approximately around 20\% $\sim$ 40\%—where the Macro-F1 score peaks as the MTR is varied. However, it is  unsound to classify a set $S$ as inconsistent only when an inconsistency of this proportion is detected.

\subsection{Fine-Tuning Efficiency}
\label{appendix: fine-tuning efficiency}

This section details our fine-tuning experiments across different domain datasets. Specifically, when fine-tuning the \scv~model—originally trained on one dataset—using N samples from a different domain, we randomly select N samples from the original dataset and combine them with N samples from the target domain for each training epoch, resulting in a total of 2N training samples per epoch. An L2 regularizer with a weight of 0.00001 was applied during fine-tuning.

\subsection{Locating Inconsistent Statements}
\label{appendix: Locating Inconsistent Statements}
In this section, we discuss the experimental setup details and the experimental results pertaining to the Locate task.

\subsubsection{Experiment Setup Detail}
\label{appendix: locate experiment setup detail}
In this section, we describe how the Locate task is performed for different model architectures.

\paragraph{LLM-based Model Architecture}  
For the LLM-based model architecture, we use chain-of-thought (CoT) prompting. The prompt is as follows:

\textbf{Prompt:}
\begin{quote}
Find the question-answer pairs among the following that are logically inconsistent with the rest. Specifically, identify the minimal collection of inconsistent pairs such that the remaining pairs are logically consistent with one another. If there are no inconsistent pairs, return nothing.

\{Few-shot CoT examples (5-shot)\}

\{Problem\}

Think step by step, and respond with your answer containing the numbers of the inconsistent pairs after the `[Inconsistent pairs]' mark.
\end{quote}

\paragraph{Binary Classifier / Energy-Based Model Architecture}  
The binary classifier and energy-based model architectures can determine the overall inconsistency of a given set $S$, but they do not directly locate which specific QA pairs are responsible for the inconsistency. Therefore, we perform the set-consistency verification repeatedly to identify the inconsistent QA pairs within $S$. In the energy-based model, lower output scores indicate that a set is more consistent. Similarly, in the binary classifier, the score for the "inconsistent" label (obtained after a softmax) is used such that lower scores indicate higher consistency.

Below, we describe the procedure for performing the Locate task using a pseudocode algorithm.

\begin{algorithm}[H]
\caption{Locate Task for Binary Classifier / Energy-Based Models}
\label{alg:locate_task}
\begin{algorithmic}[1]
\STATE \textbf{Input:} Set $S = \{s_1, s_2, \dots, s_N\}$ of QA pairs.
\WHILE {True}
    \STATE Compute the set-consistency verification result for $S$.
    \IF {$S$ is classified as \textbf{consistent}}
         \STATE \textbf{Return:} Terminate --- no inconsistent pair detected.
    \ELSIF {$S$ is classified as \textbf{inconsistent} \textbf{and} $|S| = 2$}
         \STATE \textbf{Return:} Terminate --- set too small to isolate an inconsistent pair.
    \ELSE
         \FOR {each $i = 1$ to $N$}
             \STATE Generate $S_i = S \setminus \{s_i\}$.
             \STATE Compute the energy value $E_{\theta}(S_i)$ (or the corresponding inconsistent score).
         \ENDFOR
         \STATE Let $s_j$ be the element whose removal yields the lowest energy value, i.e., $j = \arg\min_{i} E_{\theta}(S_i)$.
         \STATE \textbf{Output:} Identify $s_j$ as the inconsistent QA pair.
         \STATE Update $S \gets S_{j}$.
    \ENDIF
\ENDWHILE
\end{algorithmic}
\end{algorithm}

In summary, for the binary classifier and energy-based model architectures, we repeatedly remove the QA pair whose exclusion minimizes the energy value (or inconsistent score) until the remaining set is classified as consistent. 

\subsection{Detecting Inconsistencies in LLM Outputs}
\label{appendix: external module evaluation details}

In this section, we describe the data construction and prompt design procedures used for our experiments.

\subsubsection{Data Construction}

\paragraph{Data Collection:}  
Following the data augmentation approach of \cite{asai2020logic}, we augment the WIQA dataset \cite{tandon2019wiqa}. An example of an augmented data instance is shown below:

\begin{verbatim}
{
  "paragraph": [
    "Water from oceans, lakes, rivers, swamps, and plants turns into water vapor",
    "Water vapor forms droplets in clouds",
    "Water droplets in clouds become rain or snow and fall",
    "Some water goes into the ground",
    "Some water flows down streams into rivers and oceans",
    ""
  ],
  "choices": [
    {"label": "A", "text": "more"},
    {"label": "B", "text": "less"},
    {"label": "C", "text": "no effect"}
  ],
  "qa_pairs": [
    {
      "question": "suppose there is less water on the ground happens,  how will it 
      affect less rain will fall.",
      "answer_label": "more",
      "answer_label_as_choice": "A"
    },
    {
      "question": "suppose there is more water on the ground happens, how will it 
        affect less rain will fall.",
      "answer_label": "less",
      "answer_label_as_choice": "B"
    },
    {
      "question": "suppose more tadpoles develop in eggs happens, how will it affect
      less rain will fall.",
      "answer_label": "no_effect",
      "answer_label_as_choice": "C"
    },
    
    { ... additional QA pairs ... }
  ]
}
\end{verbatim}

\paragraph{Data Cleaning:}  
Our primary objective is to verify whether an LLM provides consistent answers when performing question-answering independently on related questions. To this end, we first group related questions by:
\begin{itemize}
  \item Grouping questions that share the same paragraph.
  \item Selecting questions whose lists of words differ by at most two words.
\end{itemize}

For example, a group of related QA pairs might include:
\begin{verbatim}
{
  "question": "suppose there is more water on the ground happens, howwill it 
affect a more intense water cycle.",
  "answer_label": "more",
  "answer_label_as_choice": "A"
},
{
  "question": "suppose there is less water on the ground happens, how will it 
  affect a more intense water cycle.",
  "answer_label": "less",
  "answer_label_as_choice": "B"
},
{
  "question": "suppose there is more water on the ground happens, how will it 
  affect a less intense water cycle.",
  "answer_label": "less",
  "answer_label_as_choice": "B"
},
{
  "question": "suppose there is less water on the ground happens, how will it
  affect a less intense water cycle.",
  "answer_label": "more",
  "answer_label_as_choice": "A"
}
\end{verbatim}

If an LLM answers all these questions correctly, the corresponding QA pairs are consistent. However, consistency does not require perfect accuracy; in our augmented WIQA dataset, the answers follow one of three formats: \texttt{more}, \texttt{less}, or \texttt{no effect}. To facilitate automated consistency measurement, we discard QA pairs with the \texttt{no effect} answer and retain only those with \texttt{more} or \texttt{less}. In this setup, if an LLM answers all questions correctly or incorrectly, it is considered to have responded consistently. Only groups with at least two QA pairs are used in our experiments.

\subsubsection{Prompt Design}

\paragraph{Prompt for Question-Answering:}  
The WIQA dataset provides a paragraph as context alongside each question. We use the following prompt to obtain the LLM's prediction for each question:

\begin{quote}
\textbf{Prompt:} Given the following paragraphs, please select the correct answer for the given question. Return only the answer.\\
\textbf{Paragraphs:} \{A collection of sentences\}\\
\textbf{Choices:} more, less\\
\textbf{Question:} \{question\}\\[1ex]
Specifically, please carefully read the question. For example, assume the question is "suppose less water in the environment happens, how will it affect a less intense water cycle." If you believe that a decrease in water leads to a less intense water cycle, you should answer \texttt{more} because less water results in a "less" intense water cycle.
\end{quote}

\paragraph{Self-Check Mechanism:}  
The self-check mechanism leverages the LLM's own architecture to perform set-level verification. In this process, the LLM is prompted—using a zero-shot chain-of-thought (CoT) approach—to verify the consistency of the question–prediction pairs it generated. 
% This self-evaluation helps verify whether the LLM's responses are mutually consistent.

\end{document}